\documentclass{article}

\usepackage[nonatbib,preprint]{neurips_2024}

\usepackage[numbers]{natbib}

\usepackage{makecell}
\usepackage{amsmath}
\usepackage[utf8]{inputenc} %
\usepackage[T1]{fontenc}    %
\usepackage{hyperref}       %
\usepackage{url}            %
\usepackage{booktabs}       %
\usepackage{amsfonts}       %
\usepackage{nicefrac}       %
\usepackage{microtype}      %
\usepackage{xcolor}         %
\usepackage{pifont}
\usepackage{tablefootnote} 
\usepackage{graphicx} %
\usepackage{xspace}
\usepackage{wrapfig}
\usepackage{amsmath}
\usepackage{multirow}

\usepackage[capitalize]{cleveref}
\crefname{section}{Sec.}{Secs.}
\Crefname{section}{Section}{Sections}
\crefname{table}{Tab.}{Tabs.}
\Crefname{table}{Table}{Tables}
\crefname{figure}{Fig.}{Figs.}
\Crefname{figure}{Figure}{Figures}
\crefname{equation}{Eq.}{Eqs.}
\Crefname{equation}{Equation}{Equations}
\usepackage{amsfonts,amssymb}

\usepackage{caption}

\makeatletter
\DeclareRobustCommand\onedot{\futurelet\@let@token\@onedot}
\def\@onedot{\ifx\@let@token.\else.\null\fi\xspace}

\def\eg{\emph{e.g}\onedot} 
\def\ie{\emph{i.e}\onedot} 
 
\def\etc{\emph{etc}\onedot}

\def\Ours{\textbf{Emu3}\xspace}
\def\OursDPO{\textbf{Emu3-DPO}\xspace}

\title{\Ours: Next-Token Prediction is All You Need}

\author{Emu3 Team\thanks{See Contributions section for full author list.}, ~~BAAI~~~~~~~~ \\
\centerline{
\url{https://emu.baai.ac.cn}~~~~~~~~
}
\vspace{2em}
}

\begin{document}

\input figs/teaser.tex

\maketitle

\begin{abstract}

While next-token prediction is considered a promising path towards artificial general intelligence, it has struggled to excel in multimodal tasks, which are still dominated by diffusion models (\eg, Stable Diffusion) and compositional approaches (\eg, CLIP combined with LLMs).
In this paper, we introduce \Ours, a new suite of state-of-the-art multimodal models trained solely with next-token prediction. By tokenizing images, text, and videos into a discrete space, we train a single transformer from scratch on a mixture of multimodal sequences.
\Ours outperforms several well-established task-specific models in both generation and perception tasks, surpassing flagship models such as SDXL and LLaVA-1.6, while eliminating the need for diffusion or compositional architectures.
\Ours is also capable of generating high-fidelity video via predicting the next token in a video sequence.
We simplify complex multimodal model designs by converging on a singular focus: tokens, unlocking great potential for scaling both during training and inference.
Our results demonstrate that next-token prediction is a promising path towards building general multimodal intelligence beyond language.
We open-source key techniques and models to support further research in this direction.

\end{abstract}

\begin{figure}[t]
    \centering
    \includegraphics[width=0.95\linewidth]{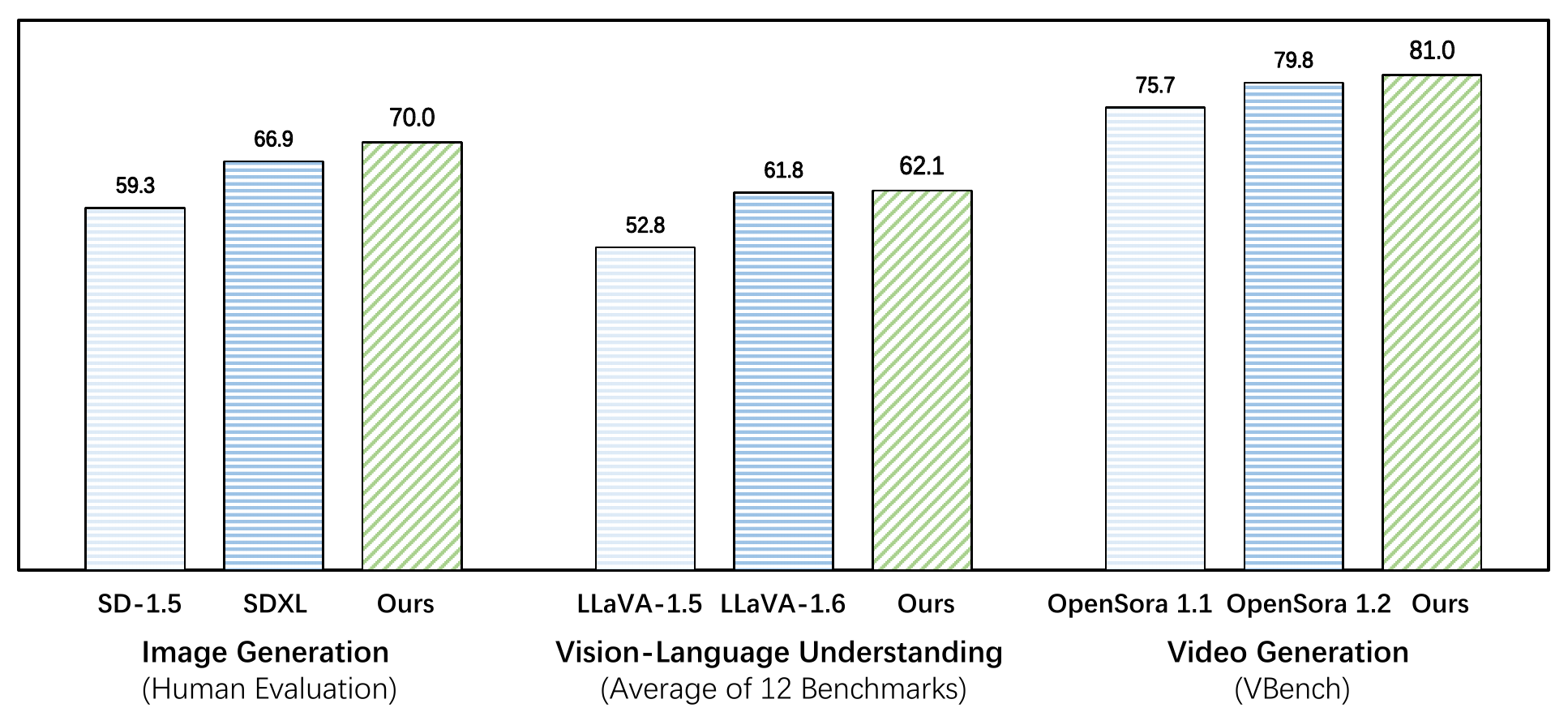}
     \caption{Comparison with open-source flagship models in vision generation and perception. Based solely on next-token prediction, \Ours beats SDXL~\cite{podell2023sdxl}, LLaVA-1.6-7B~\cite{llava-1.6}, OpenSora-1.2~\cite{opensora} respectively, dispensing with diffusion and CLIP entirely. For the image generation task, we present comparison results of human evaluation scores based on English prompts. For the vision-language understanding task, we assess the average scores across twelve benchmarks: SEEDBench-Img~\cite{datasets:seedbench}, OCRBench~\cite{liu2024hiddenmysteryocrlarge}(with normalized results), MMVet~\cite{datasets:mmvet}, POPE~\cite{datasets:pope}, VQAv2~\cite{datasets:vaqv2}, GQA~\cite{dataset:gqa}, TextVQA~\cite{datasets:textvqa}, ChartQA~\cite{masry2022chartqa}, AI2D~\cite{kembhavi2016diagram}, RealWorldQA~\cite{realworldqa}, MMMU~\cite{yue2023mmmu}, and MMbench~\cite{datasets:mmbench}. For the video generation task, we present comparison results of VBench.}
    \label{fig:overall_comparison}
    \vspace{-1em}
\end{figure}

\section{Introduction}

Next-token prediction has revolutionized the field of language models~\cite{vaswani2017attention, t5, gpt3}, enabling breakthroughs like ChatGPT~\cite{chagpt} and sparking discussions about the early signs of artificial general intelligence (AGI)~\cite{sparksofagi}. However, the applicability of this paradigm to multimodal models remains unclear, with limited evidence of its efficacy in achieving competitive performance across different tasks.

In the realm of multimodal models, vision generation has been dominated by complex diffusion models (\eg, Stable Diffusion~\cite{rombach2022high}), while vision-language perception has been led by compositional approaches such as CLIP~\cite{radford2021learning} with LLMs (\eg, LLaVA~\cite{liu2023llava}). Despite early attempts at unifying generation and perception, such as Emu~\cite{sun2023emu} and Chameleon~\cite{team2024chameleon}, these efforts either resort to connecting LLMs with diffusion models or fail to match the performance of task-specific methods tailored for generation and perception.

In this work, we present \Ours, a new set of state-of-the-art multimodal models based solely on next-token prediction, eliminating the need for diffusion or compositional approaches entirely.
We tokenize images, text, and videos into a discrete space, and jointly train a single transformer from scratch on a mix of multimodal sequences.

\Ours achieves state-of-the-art performance compared to well-established task-specific models in generation and perception tasks. 
\Ours outperforms the flagship Stable Diffusion model, \ie, SDXL~\cite{podell2023sdxl}, in both the human evaluation and the public text-to-image benchmarks such as MSCOCO-30K~\cite{chen2015microsoft}, GenEval~\cite{ghosh2024geneval}, T2I-CompBench~\cite{huang2023t2i}, and DPG-Bench~\cite{hu2024ella}.
For vision-language understanding, \Ours competes with the popular vision-language model, \ie, LLaVA-1.6~\cite{llava-1.6}, on a series of public vision-language benchmarks, including SEED-Bench~\cite{datasets:seedbench}, RealWorldQA~\cite{realworldqa}, OCRBench~\cite{liu2024hiddenmysteryocrlarge}, \etc.

\Ours is capable of generating videos.
Unlike Sora~\cite{sora} that employs the video diffusion model to generate a video from noise, \Ours simply generates a video causally by predicting the next token in a video sequence.
The model can simulate some aspects of environments, people and animals in the physical world.
With a video in context, \Ours extends the video and predicts what will happen next.
Given the user's prompt, the model can generate high-fidelity videos following the text description.
\Ours stands out and competes with other video diffusion models on the VBench benchmark~\cite{huang2024vbench} for text-to-video generation.

We open-source key techniques and models to facilitate future research in this direction. Notably, we provide a robust vision tokenizer, enabling the transformation of videos and images into discrete tokens, which was previously publicly unavailable. We also demonstrate the versatility of the next-token prediction framework, showing that direct preference optimization (DPO)~\cite{rafailov2024direct} can be seamlessly applied to autoregressive vision generation, aligning the model with human preferences.

Our results provide strong evidence that next-token prediction can serve as a powerful paradigm for multimodal models, scaling beyond language models and delivering state-of-the-art performance across multimodal tasks.
By simplifying complex model designs and focusing solely on \textit{tokens}, it unlocks significant potential for \textit{scaling} both during training and inference.
We believe that next-token prediction offers a promising path towards building general multimodal intelligence.

\begin{figure}[t]
    \centering
    \includegraphics[width=0.87\linewidth]{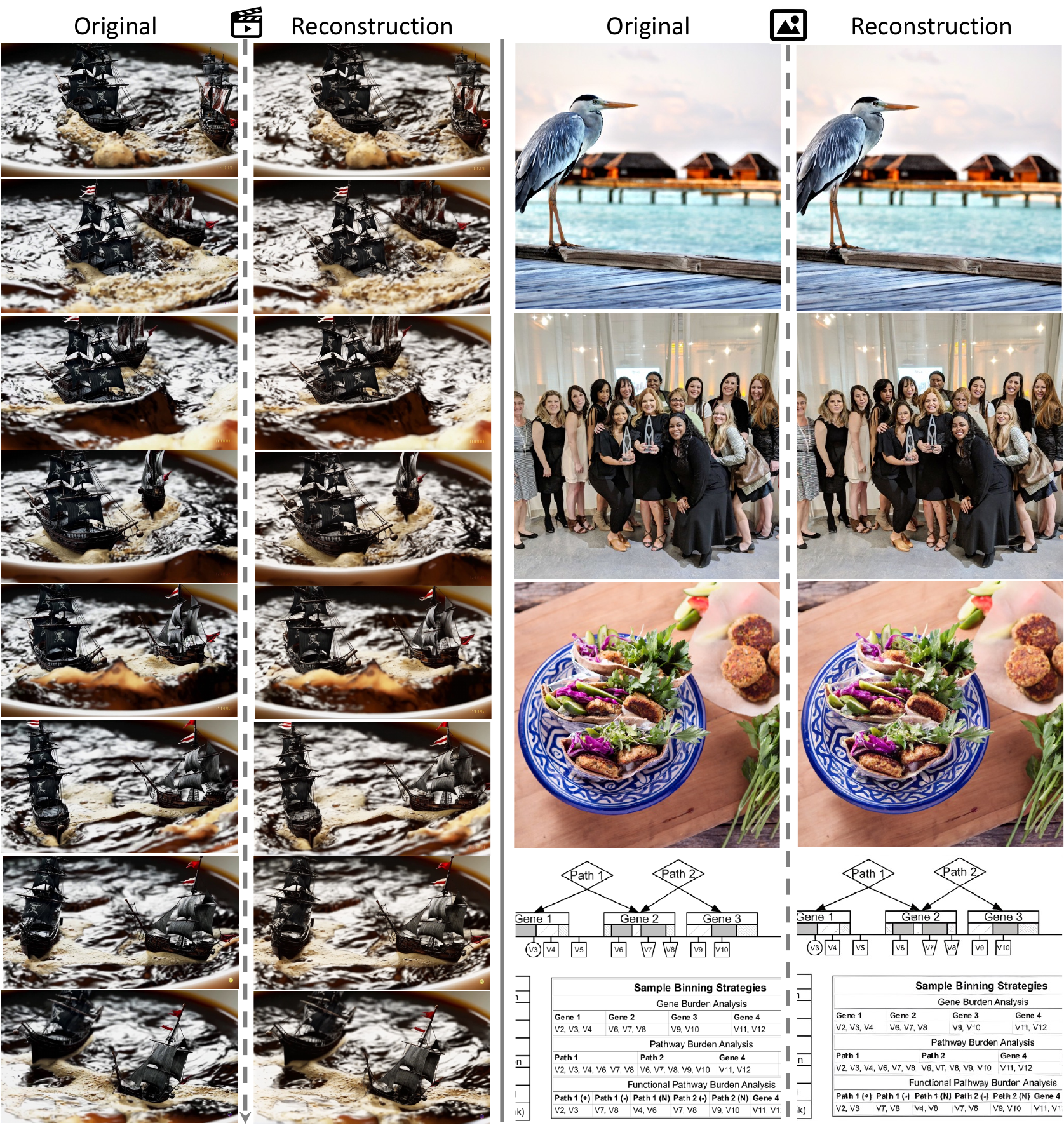}
    \caption{Reconstruction samples. Left: Original and reconstructed videos at 540 $\times$ 960 resolution, showcasing a sampling of 8 frames at 30 FPS. Right: original and reconstructed 512 $\times$ 512 resolution images. Zoom in to see the details.}
    \label{fig:recon_case}
\end{figure}

\begin{figure}[t]
    \centering
    \includegraphics[width=\linewidth]{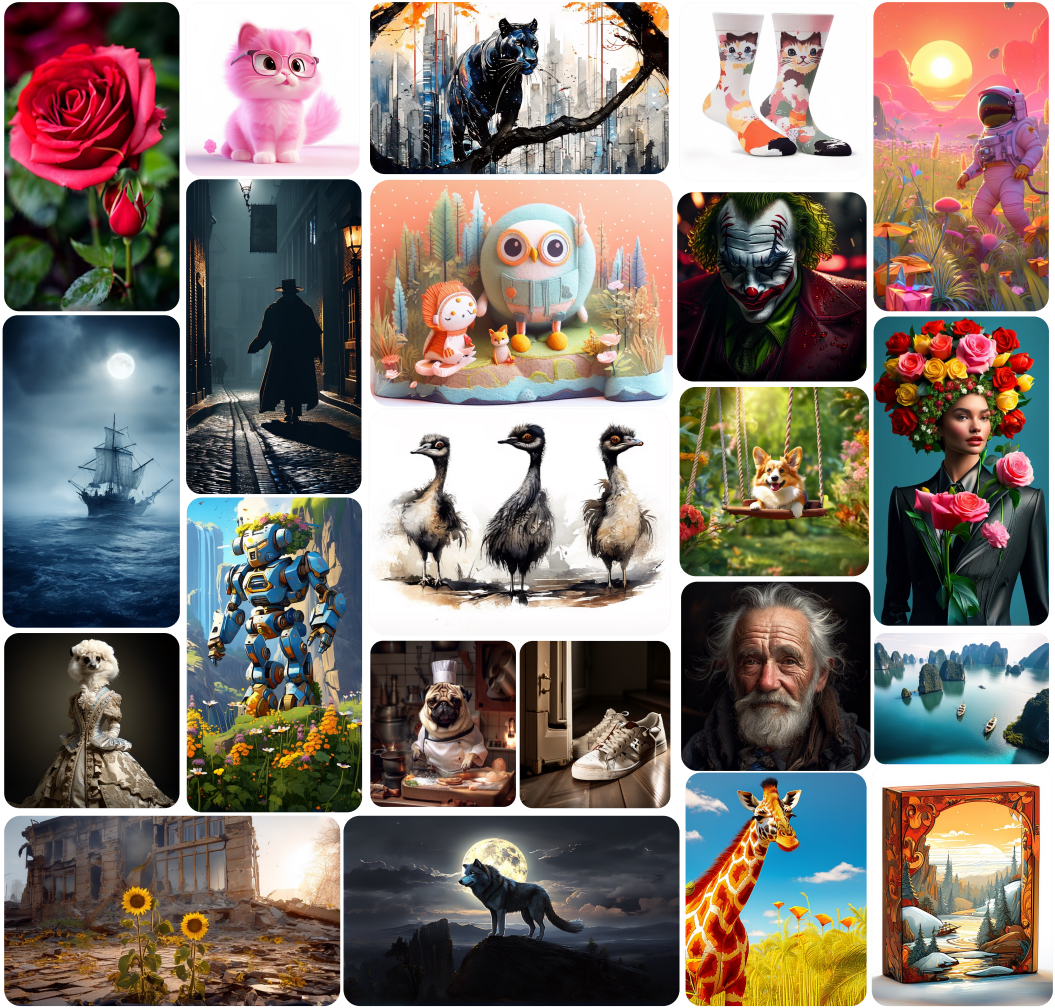}
    \caption{Qualitative results of \Ours text-to-image generation.}
    \label{fig:emu3_t2i_vis}
\end{figure}

\section{Approach}

\subsection{Data}

\Ours is trained from scratch on a mix of language, image, and video data.

\vspace{0.02\linewidth}\noindent
\textbf{Language Data.}  We use the same language data as in Aquila~\cite{zhang2024aquila2technicalreport}, which is a high-quality corpus consisting of both Chinese and English data.

\vspace{0.02\linewidth}\noindent
\textbf{Image Data.} We curate a large-scale image-text dataset comprising open-source web data, AI-generated data, and high-quality in-house data. The filtering process involves several key steps: \textbf{1}) We apply a resolution filter, discarding samples with a resolution below 512 $\times$ 512 pixels. \textbf{2}) We assess the aesthetic quality of each image using the LAION-AI aesthetic predictor\footnote{\url{https://github.com/LAION-AI/aesthetic-predictor}}, excluding images with scores below 5.5 to ensure the overall aesthetic quality. 
\textbf{3}) For images that did not pass the aesthetic filter, we employ text detection\footnote{\url{https://github.com/PaddlePaddle/PaddleOCR}\label{PaddleOCR}} and color filtering to retain non-monochromatic images and those with minimal text, improving the filtering recall of open-world images. 
\textbf{4}) Additionally, we prepare supplementary data for image understanding. By following the data processing pipeline in DenseFusion~\cite{li2024densefusion}, we extract millions of representative images that encompass a wide range of categories, including charts, table, text-rich content, and more, sourced from diverse open-source web data. 

To annotate the filtered dataset, we develop an image captioning model based on Emu2~\cite{emu2} to construct dense synthetic captions. We leverage GPT-4V~\cite{chagpt} with detailed prompts to generate approximately 1 million image-caption pairs. This annotated dataset is then used to fine-tune the Emu2-17B~\cite{emu2} model as our image captioner. Additionally, we utilize the open-source vLLM library~\cite{vllm} to accelerate the labeling process.

\vspace{0.02\linewidth}\noindent
\textbf{Video Data.}
We collect 
videos covering a wide range of categories, such as landscapes, animals, plants, games, and actions. These videos are prepossessed with a sophisticated pipeline~\cite{svd}  with the following four stages: \textbf{1}) We split the videos to scenes with PySceneDectect\footnote{\url{https://github.com/Breakthrough/PySceneDetect}}, employing both ContentDetector and ThresholdDetector to identify content changes and fade-in/out events, respectively. 
\textbf{2}) Text detection are performed using PaddleOCR\footref{PaddleOCR} and clips with excessive text coverage were removed. To reduce computational costs, we sample video frames at 2 FPS and resize the shorter edge to 256. \textbf{3}) We further calculate the optical flow~\cite{raft} to eliminate clips with minimal or extreme motion. As with the previous step, we sample and resize video frames for efficiency. The flow score is defined as the ratio between the average flow magnitude of all pixels and the shorter edge. We exclude clips with flow scores outside the acceptable range. \textbf{4}) Finally, we assess the aesthetic quality of each clip using the LAION-AI aesthetic predictor\footnotemark[1]. We sample three frames and get three scores for each clip, and clips whose lowest score is smaller than 5 are discarded.

We caption the filtered video clips using a video captioner trained based on our image captioner. The training data is initially labeled by GPT-4V~\cite{chagpt}. For each video clip, we sample eight frames and create a detailed prompt for GPT-4V to describe both the content and motion within these frames. Some of the labeled data undergoes manual revision. We then fine-tune our image captioner on this labeled data to develop our video captioner. For large-scale deployment, we accelerate captioning with vLLM~\cite{vllm}. Clips shorter than 20 seconds are captioned using 12 evenly sampled frames, while longer clips are split into 10-20 second sub-clips, each captioned independently.

\subsection{Vision Tokenizer}
We train the vision tokenizer based on SBER-MoVQGAN\footnote{\url{https://github.com/ai-forever/MoVQGAN}\label{shared-note}}, which can encode a 4 $\times$ 512 $\times$ 512 video clip or a 512 $\times$ 512 image into 4096 discrete tokens from a codebook of size 32,768. Our tokenizer achieves 4$\times$ compression in the temporal dimension and 8$\times$8 compression in the spatial dimension, applicable to any temporal and spatial resolution. Building on the MoVQGAN architecture~\cite{movqgan}, we incorporate two temporal residual layers with 3D convolution kernels into both the encoder and decoder modules to enhance video tokenization capabilities. The tokenizer is trained end-to-end on the LAION-High-Resolution\footnote{\url{https://huggingface.co/datasets/laion/laion-high-resolution}} image dataset and the InternVid~\cite{intern-vid} video dataset using combined objective functions of L2 loss, LPIPS perceptual loss~\cite{lpips}, GAN loss, and commitment loss~\cite{esser2021tamingtransformershighresolutionimage}.

Qualitative results are presented in~\cref{fig:recon_case}. We report LPIPS (computed
by the AlexNet features), PSNR, and SSIM scores in~\cref{tab:video_compression_metrics} using an evaluation dataset of 3,172 videos from Pexels\footnote{\url{https://www.pexels.com/search/videos/videos}}. The videos were reconstructed over 5 seconds while maintaining the aspect ratio. During evaluation, original and reconstructed videos were resized and cropped based on the shorter side and uniformly sampled with 8 frames at 12 FPS.
\begin{table}[t!]
    \begin{minipage}{0.495\linewidth}
        \centering
        \small
        \begin{tabular}{lc}
        \toprule 
        Configurations & VisionTokenizer \\
        \midrule
        Pretrained Weights & SBER-MoVQGAN-270M\footref{shared-note} \\
        Codebook Size & 32768 \\
        Latent Size & 4 \\
        Compression & 4 $\times$ 8 $\times$ 8 \\
        \bottomrule
        \end{tabular}
        \vspace{0.5em}
        \caption{\Ours vision tokenizer configurations.}
        \label{tab:tokenizer_configurations}
    \end{minipage}
    \hfill
    \begin{minipage}{0.5\linewidth}
        \centering
        \small
        \begin{tabular}{cccc}
        \toprule 
            Video Resolution & LPIPS$\downarrow$ & PSNR$\uparrow$ & SSIM$\uparrow$ \\ 
            \midrule
            128 $\times$ 128 & 0.099 & 21.71 & 0.630 \\
            256 $\times$ 256 & 0.109 & 21.59 & 0.622 \\
            512 $\times$ 512 & 0.112 & 22.69 & 0.690 \\
            720 $\times$ 720 & 0.110 & 24.30 & 0.771 \\
        \bottomrule
        \end{tabular}
        \vspace{0.5em}
        \caption{Video compression metrics.}
        \label{tab:video_compression_metrics}
    \end{minipage}
\vspace{-15pt}
\end{table}

\begin{wraptable}{r}{0.35\textwidth}
\vspace{-40pt}
    \centering
    \small
    \begin{tabular}{l|c}
        \toprule
        \textbf{Configurations} & \Ours \\
        \midrule
        {Parameters} & 8B \\
        {Layers} & 32 \\
        {Hidden Size} & 4096 \\
        {Intermediate Size} & 14336 \\
        {Heads} & 32 \\
        {KV Heads} & 8 \\
        {Vocabulary Size} & 184622 \\
        {RoPE Base} & 1000000 \\
        {Context Length} & 131072 \\
        \bottomrule
    \end{tabular}
    \caption{Model configurations.}
    \label{tab:model_configurations}
\vspace{-40pt}
\end{wraptable}

\subsection{Architecture}

The \Ours model retains the architectural framework of established large language models (LLMs) such as Llama-2~\cite{touvron2023llama2}, with the primary modification being the expansion of the embedding layer to accommodate discrete vision tokens. We use RMSNorm~\cite{zhang2019root} for normalization and GQA~\cite{ainslie2023gqa} for attention mechanisms, while employing the SwiGLU~\cite{shazeer2020glu} activation function and rotary positional embeddings (RoPE)~\cite{su2024roformer}. Biases in the qkv and linear projection layers are removed. Additionally, a dropout rate of 0.1 is implemented to improve training stability. We use the QwenTokenizer\footnote{\url{https://huggingface.co/Qwen/Qwen-7B/blob/main/tokenization\_qwen.py}} to tokenize multilingual texts. Detailed configurations are provided in~\cref{tab:model_configurations}.

\subsection{Pre-training}
\textbf{Data Preparation.} During pre-training we first define the multimodal data format. Unlike diffusion models that rely on an external text encoder, \Ours natively integrates textual conditional information for image/video generation. We rescale images/videos while preserving their aspect ratio to a size with an area close to 512 $\times$ 512, and then generate vision tokens using our vision tokenizer. Then, we incorporate five special tokens to merge text and vision data, creating document-like inputs for the training process. The resulting training data is structured as follows:
\[
\small{\text{[BOS] \{caption text\} [SOV] \{meta text\} [SOT] \{vision tokens\} [EOV] [EOS]}}.
\]
Where {[BOS]} and {[EOS]} are the original special tokens in the text tokenizer, {[SOV]} marking the start of the vision input, {[SOT]} marking the start of vision tokens, and {[EOV]} indicating the end of the vision input. Additionally, {[EOL]} and {[EOF]} are inserted into the vision tokens to denote line breaks and frame breaks, respectively.
The ``meta text'' contains information about the resolution for images, and for videos, it includes resolution, frame rate, and duration, all presented in plain text format.
    We also move the ``caption text'' field in a portion of the dataset to follow the [EOV] token, thereby constructing data aimed at vision understanding tasks.

\vspace{0.02\linewidth}\noindent
\textbf{Training Objective.} 
Since vision signals in \Ours are fully converted into discrete tokens, we only need to train with the next-token prediction task using the standard cross-entropy loss. To prevent vision tokens from dominating the learning process, we apply a weight of 0.5 to the loss associated with vision tokens.

\vspace{0.02\linewidth}\noindent
\textbf{Training Details.}
The \Ours model utilizes an extensive context length during pre-training to handle video data. To facilitate training, we employ a combination of tensor parallelism (TP), context parallelism (CP), and data parallelism (DP). 
We simultaneously pack text-image data into the maximum context length to fully utilize computational resources, while ensuring that complete images are not segmented during the packing process. The pre-training process is conducted in two stages. In the first stage, which does not utilize video data, training begins from scratch with a context length of 5120 for text and image data.
In the second stage, video data is introduced, and a context length of 131072 is employed.
Both stages use a learning rate of $5 \times 10^{-5}$, with a cosine annealing of the learning rate to zero.

\subsection{Post-training}

\subsubsection{Vision Generation} 
\textbf{Quality Fine-Tuning.} Following the pre-training phase, we conduct post-training for vision generation tasks to enhance the quality of generated outputs. We apply quality fine-tuning (QFT) using high-quality data. The model continues training with the next token prediction task using standard cross-entropy loss; however, supervision is applied exclusively to the vision tokens. For the image data in QFT, we select diverse high-quality sources and filtered them based on the average of three popular preference scores: HPSv2.1~\cite{wu2023human}, MPS~\cite{MPS}, and the LAION Aesthetics score~\cite{laionaesthetics}. During QFT, we increase the training data resolution from 512 pixels to 720 pixels to improve generation quality. For the video data, we sample from high-quality sources and apply stringent resolution and optical flow filters to ensure quality. Additionally, at the end of training, we use an annealing strategy to linearly decay the learning rate to zero.

\vspace{0.02\linewidth}\noindent
\textbf{Direct Preference Optimization.} Direct Preference Optimization (DPO)~\cite{rafailov2024direct}, an effective approach for better aligning models with human preferences. 
We adopt DPO techniques for autoregressive multimodal generation tasks, leveraging human preference data to enhance model performance. We divide the dataset construction into three steps: \textbf{1}) We perform 8-10 inferences for each user-collected prompt (\( p \)) using the quality fine-tuned model, creating an initial data pool (\( x \)). \textbf{2}) Each prompt is evaluated by three voters, focusing on vision appeal and prompt alignment. \textbf{3}) Based on the scores, the highest scoring sample is chosen, and the lowest is rejected to form a triplet (\(p_i\), \(x^{chosen}_i\), \(x^{rejected}_i\)) with the prompt for further training.
Specifically, the tokens from the data construction process are stored for direct use in future training phases. This strategy eliminates reconstruction differences caused by re-tokenization. \OursDPO minimizes the DPO loss and the next token prediction cross-entropy loss to fine-tune the QFT model.

\subsubsection{Vision-Language Understanding}\label{sec:vis_understand}
The pretrained model undergoes a two-stage post-training process for vision-language understanding: 1) image-to-text training, and 2) instruction tuning. During the first stage, our approach integrates image understanding data with pure-language data, while losses associated with vision tokens are disregarded for text-only prediction. Each image is resized to a resolution of about 512 $\times$ 512 while preserving the original aspect ratio. In the second stage, we sample a subset of question-answer pairs from ~\cite{llava-onevision} to enhance the vision instruction following ability.
Images below 512 $\times$ 512 or above 1024 $\times$ 1024 will be resized to the lower or upper resolution limit while keeping the aspect ratio accordingly, while others maintain their original resolution.

\section{Main Results}

\subsection{Image Generation}

\begin{figure}[t]
    \centering
    \begin{minipage}{0.64\linewidth}
        \centering
        \includegraphics[width=\linewidth]{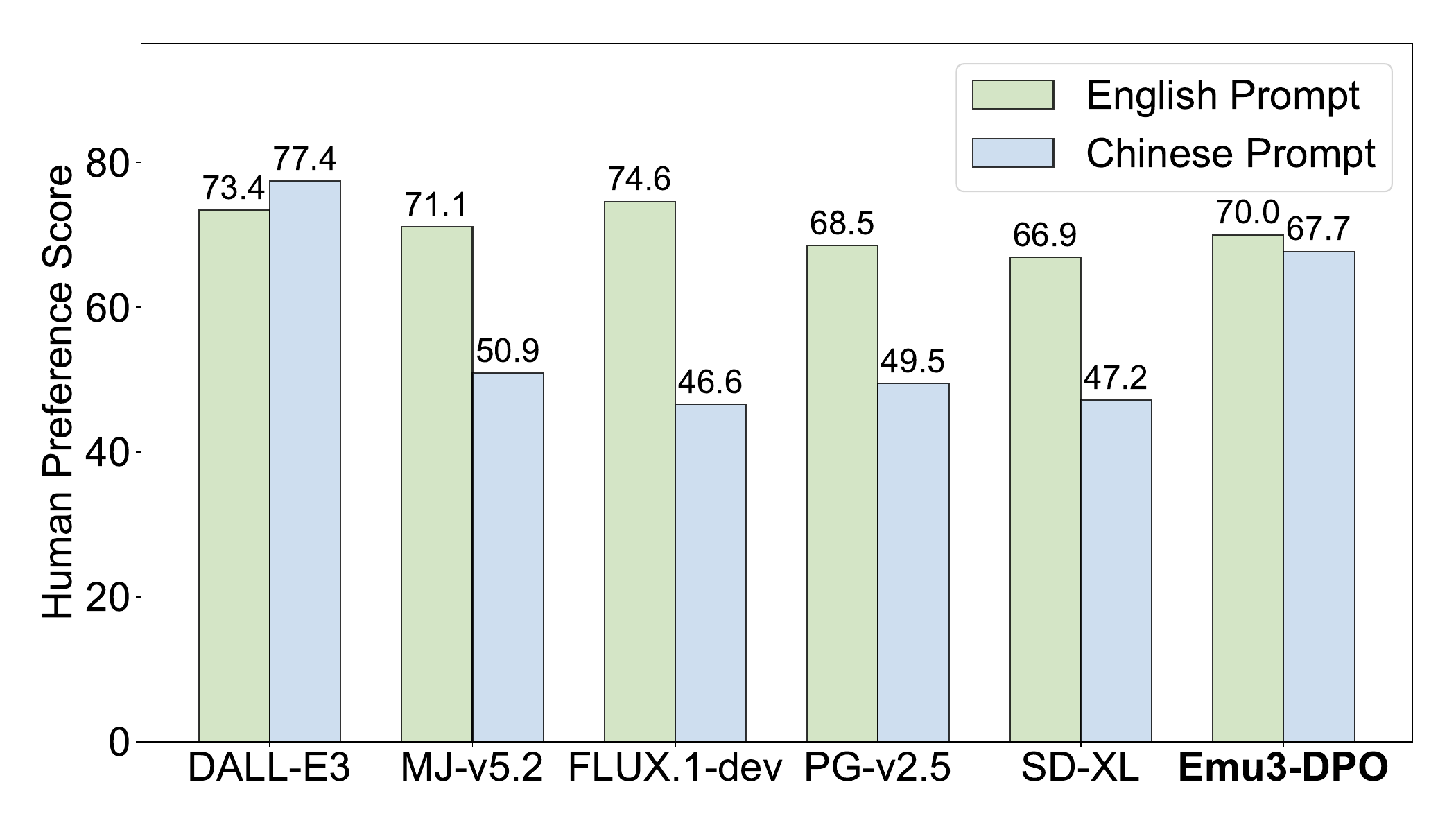}
        \caption{Human evaluation overall score comparison of closed and open generative image models under English and Chinese prompts.}
        \label{fig:human_preference_comparison}
    \end{minipage}
    \hfill
    \begin{minipage}{0.34\linewidth}
        \centering
        \includegraphics[width=\linewidth]{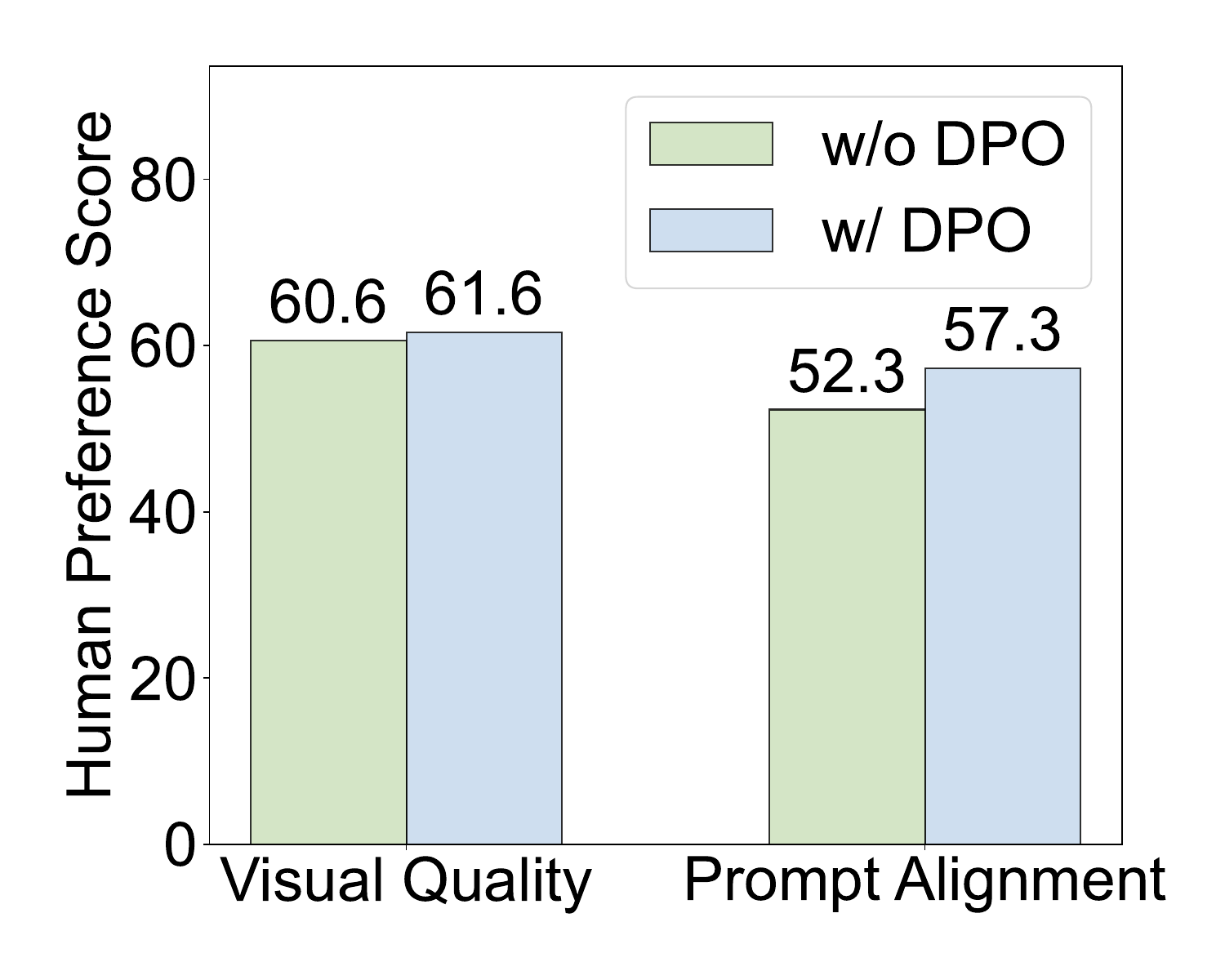}
        \caption{DPO improves visual quality and prompt alignment.}
        \label{fig:dpo_vs_qft}
    \end{minipage}
\end{figure}

\subsubsection{Automated Metric Evaluation}

We present the performance of \Ours through automated metric evaluation on popular text-to-image benchmarks: MSCOCO-30K~\cite{chen2015microsoft}, GenEval~\cite{ghosh2024geneval}, T2I-CompBench~\cite{huang2023t2i}, and DPG-Bench~\cite{hu2024ella}. The comparison results of \Ours against diffusion methods, autoregressive diffusion methods, and autoregressive-based methods across these four benchmarks are shown in \cref{tab:text2image_fid}. Our method outperforms autoregressive diffusion methods in image-text alignment evaluation and is comparable to state-of-the-art diffusion-based models, despite not utilizing any pre-trained language models. 

We report the results of GenEval and T2I-CompBench after employing a rewriter to expand short prompts. Due to \Ours utilizing a significant proportion of synthetic labels during training, it exhibits superior performance in dense captioning compared to shorter prompts. However, the evaluation prompts in GenEval and T2I-CompBench are too brief to accurately reflect the model's true performance. Following DALL-E 3, we also report our evaluation results using GPT-4V as the rewriter. The GenEval overall score results indicate that \Ours significantly outperforms Chameleon, a multi-modal autoregressive model, as well as the latest autoregressive diffusion methods, Show-O and Transfusion. Additionally, \Ours surpasses SDXL and matches the performance of state-of-the-art diffusion models, including DALL-E 3. Detailed comparisons across all dimensions, including results from the original prompts, are provided in Appendix~\ref{sec:Image-eval}.

To further assess state-of-the-art text-to-image methods, particularly diffusion models, we evaluate the alignment between generated images and text conditions using T2I-CompBench.
\Ours demonstrates competitive performance compared to SoTA diffusion-based models. Additionally, we compare our models with state-of-the-art (SoTA) models on the DPG-Bench, which features longer prompts with more detailed information for evaluation. Our \OursDPO achieves an overall score of 81.6, surpassing SDXL and PixArt-alpha, and is comparable to DALL-E 3, providing further evidence of the model's ability to follow long prompts. When comparing \Ours with \OursDPO, we observe a slight decline in the evaluation results after applying DPO, which may be attributed to preferences in our DPO datasets that emphasize overall aesthetic quality--a focus that differs from the domains of the automated evaluation models, complicating conclusions drawn solely through automated evaluation. We therefore introduced human evaluation in Sec.\ref{sec:he}.

\begin{table}[t]
    \centering
    \resizebox{\linewidth}{!}{
        \begin{tabular}{l|c|ccc|c|ccc|c}
            \toprule
                    &    & \multicolumn{3}{c}{MSCOCO}  & \multicolumn{1}{c}{GenEval}  &  \multicolumn{3}{c}{T2I-CompBench}   &  \multicolumn{1}{c}{DPG-Bench}  \\
            Method & Text Pretrain & CLIP-I & CLIP-T & FID       & Overall          &  Color  & Shape & Texture      &  Average   \\
            \midrule
            \multicolumn{9}{l}{\textit{Diffusion-based}} \\
            \midrule
            SDv1.5~\cite{rombach2022high}           &  CLIP ViT-L/14    & 0.667 & 0.302 & 9.93 & 0.43 & 0.3730 & 0.3646 & 0.4219 & 63.18   \\
            DALL-E 2  ~\cite{ramesh2022hierarchical} & CLIP ViT-H/16 & - & 0.314 & 10.93 & 0.52 & 0.5750 & 0.5464 & 0.6374 & -   \\
            SDv2.1~\cite{rombach2022high}           &  CLIP ViT-H/14    & - & - & - & 0.50 & 0.5694 & 0.4495 &  0.4982 & -   \\
            SDXL~\cite{podell2023sdxl}              &  CLIP ViT-bigG   & 0.674 & 0.310  & - & 0.55 & 0.6369 & 0.5408 & 0.5637 & 74.65 \\
            PixArt-alpha~\cite{chen2023pixart}  &  Flan-T5-XXL   & - & - & 7.32 & 0.48 & 0.6886 & 0.5582 & 0.7044 & 71.11   \\
            DALL-E 3 ~\cite{betker2023dalle3}       & Flan-T5-XXL    & -  & 0.320 & -  &~~0.67\dag  & 0.8110 & 0.6750 & 0.8070 & 83.50    \\ 
            SD3~\cite{esser2024scaling} &  Flan-T5-XXL   & - & - & - & 0.74 & - & - & - & -  \\

            \midrule
            \multicolumn{9}{l}{\textit{Autoregressive meets diffusion}} \\
            \midrule
            Emu~\cite{sun2023emu}    &  LLaMA-7B   & 0.656 & 0.286 & 11.6  & - & - & - & - & -  \\
            Show-o~\cite{show-o}      & Phi-1.5 & - & - & 9.24 & 0.53 & - & - & - & - \\
            Transfusion~\cite{zhou2024transfusion} & - & - & - & 6.78 & 0.63 & - & - & - &  - \\

            \midrule
            \multicolumn{9}{l}{\textit{Autoregressive-based}} \\
            \midrule
            Chameleon~\cite{team2024chameleon}   & - & - & - & 26.74 & 0.39 & - & - & - & - \\
            LlamaGen~\cite{sun2024autoregressive}   & FLAN-T5 XL & - & - & - & 0.32  & - & - & - & - \\
            \Ours                       & - & 0.689 & 0.313 & 12.8  &~~0.66\dag & 0.7913\dag & 0.5846\dag & 0.7422\dag & 80.60  \\
            \OursDPO                                    & - & 0.680 & 0.312 & 19.3 &~~0.64\dag & 0.7544\dag & 0.5706\dag & 0.7164\dag & 81.60 \\
            \bottomrule
        \end{tabular}
    }
    \vspace{0.5em}
    \caption{\textbf{Comparison with state-of-the-art models on text-to-image benchmarks.} We evaluate on MSCOCO-30K~\cite{chen2015microsoft}; GenEval~\cite{ghosh2024geneval}; T2I-CompBench~\cite{huang2023t2i} and DPG-Bench~\cite{hu2024ella}. \dag ~result is with rewriting.}
    \label{tab:text2image_fid}
\end{table}

\subsubsection{Human Evaluation}
\label{sec:he}
We conduct a human evaluation comparing the text-to-image generation capabilities of different models. A set of 100 diverse user prompts is collected, and each is evaluated by three independent voters. The evaluation focuses on two main aspects: visual quality and prompt following, with a weighted score reflecting the overall performance. As shown in Fig.\ref{fig:human_preference_comparison}, we present a comparison of human preferences for current closed and open generative image models. The results indicate that \Ours outperforms SDXL and is on par with DALL-E 3 and MJ-v5.2 in terms of overall score. Furthermore, Fig.~\ref{fig:dpo_vs_qft} demonstrates the impact of alignment through DPO fine-tuning, which effectively improves visual quality and prompt following.

\subsubsection{Qualitative Results} 
Fig.~\ref{fig:emu3_t2i_vis} shows 25 images generated by \Ours to showcase its capabilities. \Ours supports flexible resolutions, aspect ratios, and is capable of handling various styles.

\begin{figure}[htbp]
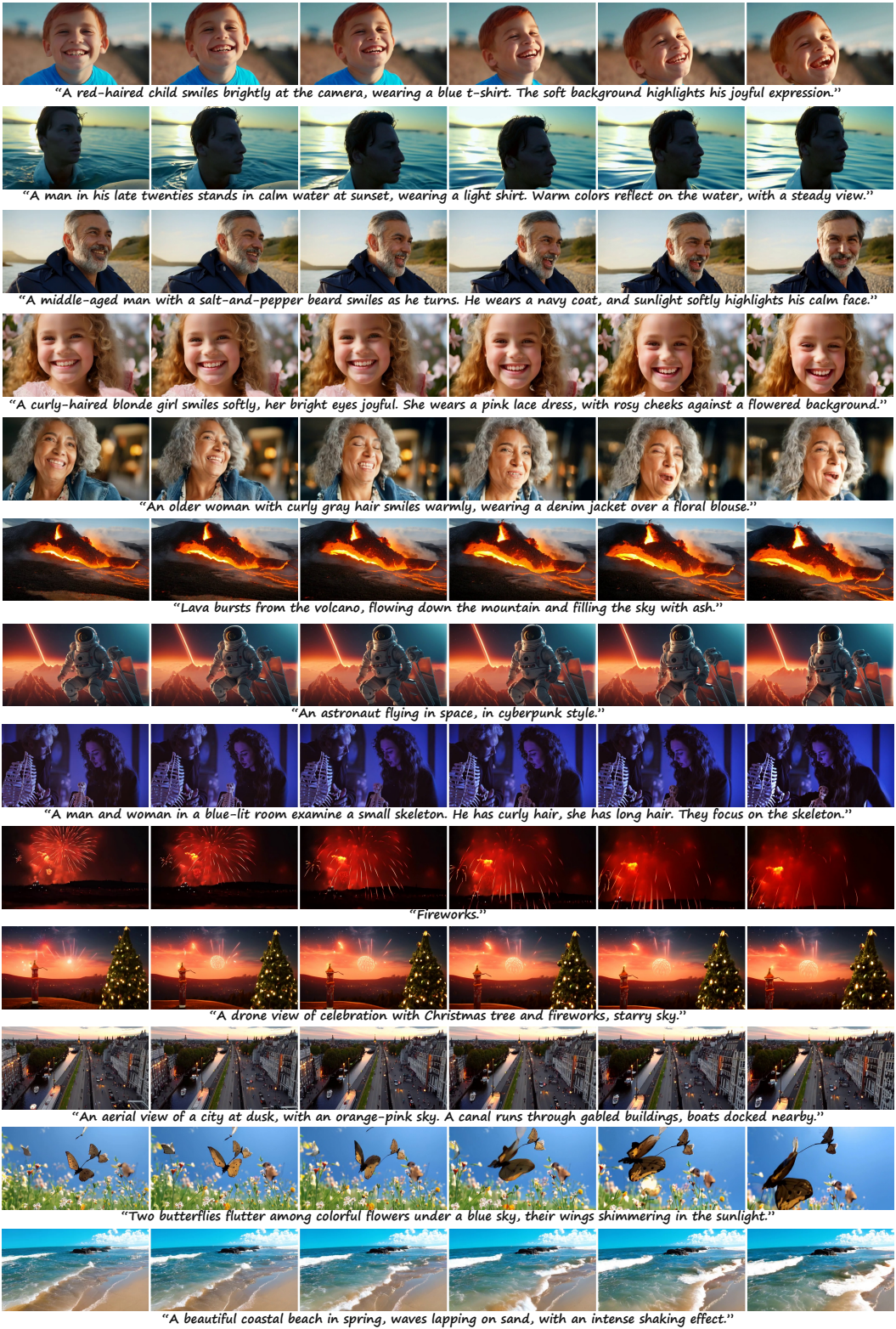

    \centering
    \includegraphics[width=1.0\linewidth, trim=0cm 3cm 0cm 0cm, clip]{figs/t2vcases_with_prompt_1.pdf}
    \includegraphics[width=1.0\linewidth]{figs/t2vcases_with_prompt_2.pdf}
    \caption{Qualitative results of \Ours text-to-video generation.}
    \label{fig:emu3_t2v_vis}
\end{figure}

\subsection{Video Generation}

Consistent with training stage, \Ours natively supports the generation of 5-second videos at 24 FPS and can be infinitely extended through an autoregressive approach. Fig.~\ref{fig:emu3_t2v_vis} presents qualitative examples of video generation, with 6 frames extracted from the first 3 seconds for showcase.

 We conducted a quantitative comparison between \Ours and the 13 best-performing open-source and proprietary text-to-video models. The used benchmark is VBench~\cite{huang2024vbench}, a comprehensive toolkit for evaluating video generation performance, which assesses the quality and semantic capabilities of each model across 16 dimensions. 
Aside from \Ours, which is an autoregressive model, all other publicly comparable methods are diffusion models.
Nevertheless, as shown in Tab.~\ref{tab:comp_video}., \Ours demonstrates highly competitive results compared to other state-of-the-art models in the overall score. Specifically, while it falls short of the most advanced proprietary models such as Kling~\cite{kling} and Gen-3~\cite{gen3}, it outperforms the majority of open-source text-to-video models. These results highlight the strong video generation capabilities of \Ours.
 
\begin{table}[t]
    \centering
    \setlength{\tabcolsep}{0.095cm}
\resizebox{\linewidth}{!}{
\begin{tabular}{l >{\centering\arraybackslash}p{1.2cm} ccccccccccccc}
\toprule
          Models &  Type  & \makecell{Total\\score} & \makecell{Motion\\smoothness} & \makecell{Dynamic\\degree} & \makecell{Aesthetic\\quality} & \makecell{Object\\class} & \makecell{Multiple\\objects} & \makecell{Human\\action} & \makecell{Spatial\\relationship} &  \makecell{Scene} & \makecell{Appearance\\style} & \makecell{Subject\\consistency} & \makecell{Background\\consistency} \\

\midrule
                 ModelScope~\cite{wang2023modelscope} &  Diff &      75.75 &             95.79 &          66.39 &             56.39 &        82.25 &            38.98 &         92.4 &                33.68 &  39.26 &               25.67 & 89.87 & 95.29 \\
                      LaVie~\cite{wang2023lavie} & Diff &       77.08 &             96.38 &          49.72 &             54.94 &        91.82 &            33.32 &         96.8 &                34.09 &  52.69 &            23.56 &  91.41 & 97.47 \\
          OpenSoraPlan V1.1~\cite{opensoraplan} &   Diff &     78.00 &             98.28 &          47.72 &             56.85 &         76.3 &            40.35 &         86.80 &                53.11 &  27.17 &             22.90 &  95.73 & 96.73 \\
                     Show-1~\cite{zhang2023show} & Diff &        78.93 &             98.24 &          44.44 &             57.35 &        93.07 &            45.47 &         95.60 &                 53.50 &  47.03 &            23.06 &     95.53 &  98.02 \\
         OpenSora V1.2~\cite{opensora} &  Diff &       79.76 &              98.50 &          42.39 &             56.85 &        82.22 &            51.83 &         91.20 &                68.56 &  42.44 &            23.95 &  96.75 & 97.61 \\
             AnimateDiff-V2~\cite{guo2023animatediff} &  Diff &       80.27 &             97.76 &          40.83 &             67.16 &         90.90 &            36.88 &         92.60 &                 34.60 &  50.19 &            22.42  & 95.30 & 97.68 \\
            Gen-2~\cite{gen2} & Diff &        80.58 &             99.58 &          18.89 &             66.96 &        90.92 &            55.47 &         89.20 &                66.91 &  48.91 &            19.34 &  97.61 & 97.61 \\
         Pika~\cite{pika} & Diff &       80.69 &              99.50 &           47.50 &             62.04 &        88.72 &            43.08 &         86.20 &                61.03 &  49.83 &            22.26 &   96.94 & 97.36  \\
           VideoCrafter-2.0~\cite{chen2024videocrafter2} & Diff &        80.44 &             97.73 &           42.50 &             63.13 &        92.55 &            40.66 &           95.00 &                35.86 &  55.29 &            25.13 &  96.85  & 98.22 \\
            T2V-Turbo (VC2)~\cite{li2024t2v} & Diff &        81.01 &             97.34 &          49.17 &             63.04 &        93.96 &            54.65 &         95.20 &                38.67 &  55.58 &            24.42 &    96.28 & 97.02 \\
 CogVideoX-5B~\cite{yang2024cogvideox} &  Diff &       81.61 &             96.92 &          70.97 &             61.98 &        85.23 &            62.11 &         99.40 &                66.35 &   53.20 &            24.91 &   96.23 &  96.52 \\
Kling (2024-07)~\cite{kling} &  Diff &      81.85 &              99.40 &          46.94 &             61.21 &        87.24 &            68.05 &         93.40 &                73.03 &  50.86 &            19.62 &   98.33 & 97.60 \\
            Gen-3~\cite{gen3} & Diff &       82.32 &             99.23 &          60.14 &             63.34 &        87.81 &            53.64 &         96.4 &                65.09 &  54.57 &            24.31 &  97.10 & 96.62 \\
                       \Ours (Ours) & AR &        80.96 &             98.93&           79.27 &             59.64 &        86.17 &            44.64 &        77.71 &                68.73 &   37.11 &            20.92 &  95.32 & 97.69 \\
\bottomrule
\end{tabular}
}
\vspace{0.5em}
\caption{\textbf{Comparison with state-of-the-art text-to-video models on VBench~\cite{huang2024vbench} benchmark.}. We selected 11 out of the 16 evaluation dimensions from VBench, along with the final score, for presentation. Except for Emu3, which is an autoregressive (AR) model, \textit{all other publicly comparable method are diffusion (Diff) models}. The higher metrics indicate the better results.} 
    \label{tab:comp_video}
\end{table}

\subsection{Future Prediction}

\begin{figure}[t]
    \centering
    \includegraphics[width=1.0\linewidth]{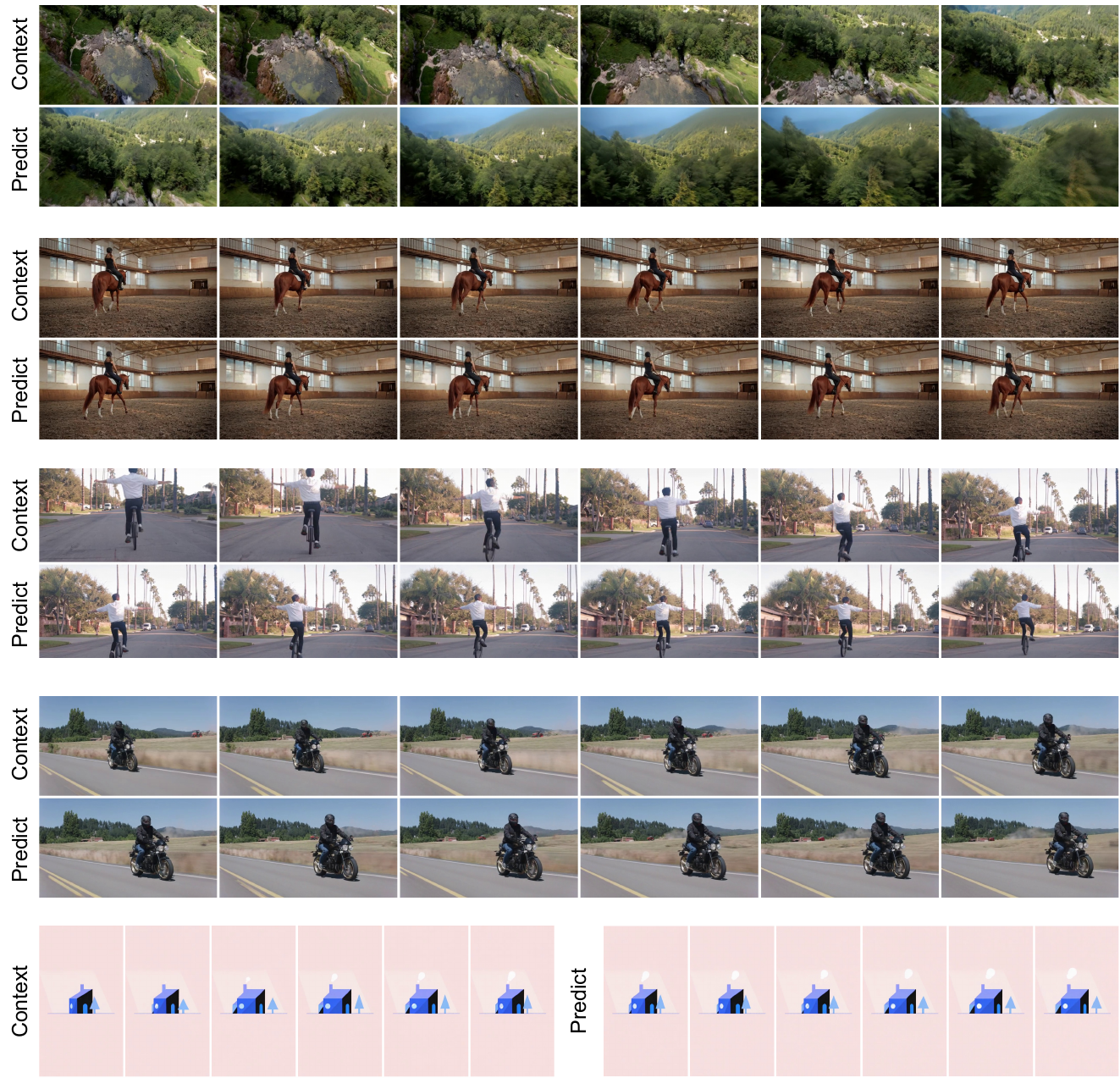}
    \caption{Qualitative results of \Ours on video extension. We sample 3 frames per second for display.} 
    \label{fig:future_prediction}
\end{figure}

\Ours can extend videos by predicting future frames. In Fig.~\ref{fig:future_prediction}, we illustrate qualitative examples of video extension, where 2-second videos at 24 FPS are tokenized into discrete vision tokens as context. \Ours predicts the subsequent 2 seconds of content in the same form of discrete vision tokens, which can be detokenized to generate future predicted videos. These examples demonstrate that utilizing only next-token prediction facilitates the temporal extension of videos, including the prediction of human and animal actions, interactions with the real world, and variations in three-dimensional animations. Furthermore, by extending the video duration in this manner, our approach is capable of iteratively generating videos that surpass its contextual length. We have observed that successfully expanding future video frames by 8 seconds using 2 seconds of video data as context is feasible.

\begin{table}[h]
    \centering
    \setlength{\tabcolsep}{0.075cm}
    \resizebox{\linewidth}{!}{
        \begin{tabular}{ll cccc cccc cccc ccc}
            \toprule
            Method & Pretrained-LLM
            & SEEDB & OCRB & MMV & POPE
            & VQAv2 & GQA & SQA & TQA 
            & CQA & DVQA & IVQA & AI2D 
            & RWQA & MMMU & MMB
            \\
            \midrule
            \multicolumn{11}{l}{\textit{Encoder-based}}\\
            \midrule
            InstructBLIP~\cite{dai2023instructblip}
            & Vicuna-7B
            & 53.4 & 276  & 26.2 & -- 
            & --   & 49.2 & 60.5 & 50.1 
            & 12.5 & 13.9 & -- 
            & 33.8 & 37.4 & 30.6 & 36.0 
            \\
            IDEFICS-9B~\cite{idefics}
            & LLaMA-7B
            & --   & 252  & --   & -- 
            & 50.9 & 38.4 & --   & 25.9 
            & --   & --   & -- 
            & 42.2 & 42.1 & 18.4 & 48.2 
            \\
            QwenVL-Chat~\cite{qwen}
            & Qwen-7B
            & 58.2 & 488  & --   & -- 
            & ~78.2{*} & ~57.5{*} & 68.2 & 61.5 
            & 49.8 & 66.3 & -- 
            & 45.9 & 49.3 & 35.9 & 60.6 
            \\
            LLaVA-1.5~\cite{llava-1.5}
            & Vicuna-7B
            & 64.3 & 318 & 30.5 & 85.9
            & ~78.5{*} & ~62.0{*} & 66.8 & 46.1 & 18.2 & 28.1 & 25.8
            & 54.8 & 54.8 & 35.3 & 64.3 
            \\
            InternVL-Chat~\cite{internvl-1.5} 
            & Vicuna-7B
            & --  & --    & --  & 86.4
            & ~79.3{*} & ~62.9{*} & --  & 57.0 
            & --   & --  & -- 
            & --   & --  & -- & --   
            \\
            mPLUG-Owl2~\cite{mplug-owl2}
            &LLaMA2-7B
            & 57.8 & 255  & 36.5 & 86.2
            & ~79.4{*} & ~56.1{*} & 68.7 & 58.2 
            & 22.8 & --   & --  
            & 55.7 & 50.3 & 32.7 & 64.5 
            \\
            ShareGPT4V~\cite{sharegpt4v}
            & Vicuna-7B
            & --  & 371   & 37.6 & --
            & ~80.6{*} & ~63.3{*} & 68.4 & 60.4 
            & 21.3 & --   & --  
            & 58.0 & 54.9 & 37.2 & 68.8 
            \\
            LLaVA-1.6(HD)~\cite{llava-1.6}
            &Vicuna-7B
            & 64.7 & 532  & 43.9 & 86.5	
            & ~81.8{*} & ~64.2{*} & 70.2 & 64.9 
            & ~54.8{*} & ~74.4{*} & 37.1 
            & ~66.6{*} & 57.8 & 35.1 & 67.4 
            \\
            VILA~\cite{lin2024vila}
            &LLaMA2-7B
            & 61.1 & --   & 34.9 & 85.5
            & ~80.8{*} & ~63.3{*} & 73.7 & 66.6 
            & --   & --   & -- 
            & --   & --   & -- & 68.9 
            \\
            \midrule
            \multicolumn{11}{l}{\textit{Encoder-free}}
            \\
            \midrule
            Fuyu-8B(HD)~\cite{fuyu-8b}
            & Persimmon-8B
            & --   & --   & 21.4 & 74.1
            & 74.2 & --   & --   & -- 
            & --   & --   & -- 
            & 64.5 & --   & 27.9  & 10.7 
            \\
            Chameleon-MT-34B~\cite{team2024chameleon}
            &--
            & --   & --   & --   & --
            & 69.6 & --   & --   & -- 
            & --   & --   & -- 
            & --   & --   & -- & --  
            \\
            Show-o~\cite{show-o} 
            &Phi-1.5-1.3B
             & -- & --     & --   & 73.8
            & ~59.3{*} & ~48.7{*} & --   & -- 
            & --   & --   & -- 
            & --   & --   & 25.1 & --  
            \\
            EVE-7B(HD)~\cite{eve} 
            &Vicuna-7B
            & 56.8 & --   & 25.7 & 85.0 & ~78.6{*} 
            & ~62.6{*} & 64.9 & 56.8 
            & --   & --   & -- 
            & --   & --   & -- & 52.3
            \\
            \Ours
            & -- 
            & 68.2  & 687      & 37.2     & 85.2
            & ~75.1{*} & ~60.3{*} & ~89.2{*} & 64.7 
            & ~68.6{*} & ~76.3{*} & ~43.8{*} 
            & ~70.0{*} & 57.4 & 31.6 & 58.5
            
            \\
            \bottomrule
            \\
            \end{tabular}
        }
        \caption{\textbf{Comparison on vision-language benchmarks.} We collect evaluations including:  
            SEEDB: SEEDBench-Img~\cite{datasets:seedbench};
            OCRB: OCRBench~\cite{liu2024hiddenmysteryocrlarge};
            MMV: MMVet~\cite{datasets:mmvet};
            POPE~\cite{datasets:pope}; 
            VQAv2~\cite{datasets:vaqv2}; 
            GQA~\cite{dataset:gqa};  
            SQA: ScienceQA-Img~\cite{datasets:scienceqa}; 
            TVQA: TextVQA~\cite{datasets:textvqa};   
            CQA: ChartQA~\cite{masry2022chartqa};
            DVQA: DocVQA~\cite{mathew2021docvqa};
            IVQA: InfoVQA~\cite{mathew2022infographicvqa};
            AI2D~\cite{kembhavi2016diagram};
            RWQA: RealWorldQA~\cite{realworldqa};
            MMMU~\cite{yue2023mmmu};
            MMB: MMBench~\cite{datasets:mmbench}.
        * The images of related training datasets are observed during training.
        }
        \label{tab:v2t_results}
\end{table}

\subsection{Vision-Language Understanding}

To evaluate the vision-language understanding capabilities of \Ours fine-tuned in \cref{sec:vis_understand}, we test our model across various public vision-language benchmarks. The primary results, detailed in \cref{tab:v2t_results}, compare two categories of methods: \textbf{1}) encoder-based approaches that utilize pretrained CLIP vision encoders, and \textbf{2}) encoder-free methodologies that operate without pretrained encoders. \Ours stands out as a pure encoder-free method, notably surpassing its counterparts across several benchmarks. This achievement is made without depending on a specialized pretrained LLM and CLIP, underscoring intrinsic capabilities and promising potential of \Ours in multimodal understanding.

\section{Related Work}

\textbf{Vision-Language Understanding.} CLIP \cite{radford2021learning} learns generalizable vision representations through contrastive learning on massive image-text pairs, achieving impressive zero-shot results in image classification tasks. Flamingo \cite{alayrac2022flamingo}, by connecting pretrained language models and vision encoders akin to CLIP, initially showcases promising few-shot multimodal understanding capabilities. The increasing availability and progress of LLMs have popularized the fusion of pretrained vision encoders with LLMs, forming a common approach to train extensive vision-language models (VLMs).
The BLIP series \cite{li2022blip,li2023blip}, MiniGPT4 \cite{zhu2023minigpt}, and LLaVA \cite{liu2023llava} exhibit encouraging results by linking vision encoders with LLMs and training on image-text pairs and vision instruction tuning data. Further improvements in performance are seen in LLaVA series \cite{llava-1.5, llava-1.6} and other impressive works \cite{qwen, internvl} through curated datasets and improved training strategies. While models like Fuyu \cite{fuyu-8b} and EVE \cite{eve} introduce encoder-free vision-language architectures that feed image patches into LLMs, they still face challenges in competing with state-of-the-art VLMs.
For the first time, \Ours demonstrates that a decoder-only model trained solely on next-token prediction can achieve comparable or even superior performance compared to encoder-based VLMs. This paves the way for further improvement of such architecture.

\textbf{Vision Generation.} Recent advancements in vision generation have been largely dominated by diffusion models~\cite{rombach2022high, ramesh2022hierarchical, podell2023sdxl, Peebles2022DiT, betker2023dalle3}. 
These models demonstrate impressive capabilities in generating high-resolution images via the diffusion process.
The open-source release of the Stable Diffusion series has led to widespread research and development in this direction.
Another research line is to train autoregressive models to generate images via predicting the next token in a sequence, such as DALL-E~\cite{ramesh2021zeroshot}, CogView~\cite{ding2021cogview}, and Parti~\cite{yu2022scaling}.
VideoGPT~\cite{yan2021videogpt} and VideoPoet~\cite{kondratyuk2023videopoet} also leverage autoregressive approaches in the video domain. 
However, they either fail to match the performance with diffusion models or rely on cascade/compositioinal approaches, \eg, VideoPoet uses a two-stage generate-and-refine framework and an extra text encoder.
In this work, \Ours demonstrates state-of-the-art image and video generation capabilities with a single Transformer decoder.
Notably, we open source to support further research and development in this direction.

\textbf{Unified Understanding and Generation.} There have been early efforts to unify vision understanding and generation~\cite{sun2023emu, yu2023scaling, ge2024seed, dong2023dreamllm}, exploring various generative objectives on image and text data. Emu and Emu2~\cite{sun2023emu, emu2} introduce a unified autoregressive objective: predicting the next multimodal element, by regressing visual embeddings or classifying textual tokens. CM3Leon~\cite{yu2023scaling} and Chameleon~\cite{team2024chameleon} trained token-based autoregressive models on mixed image and text data.
More recent methods like TransFusion~\cite{zhou2024transfusion} and Show-o~\cite{show-o} attempt to combine diffusion and autoregressive approaches to boost performance.
However, these models still fall behind task-specific architectures like SDXL~\cite{podell2023sdxl} and LLaVA-1.6~\cite{llava-1.6} in terms of vision generation and understanding. 
\Ours for the first time demonstrates that next-token prediction across images, video, and text can surpass these well-established models, without relying on compositional methods.

\section{Conclusion}
In this paper, we introduced \Ours, a new series of multimodal models that excel at multimodal generation and perception through next-token prediction. 
By tokenizing images, text, and videos into a discrete space and training a single transformer from scratch, \Ours not only eliminates the reliance on diffusion and compositional methods but also surpasses the performance of established task-specific models such as SDXL and LLaVA-1.6.
Our results provide compelling evidence that next-token prediction can serve as a powerful paradigm for multimodal models, scaling beyond language models and delivering state-of-the-art performance across diverse tasks, including challenging video generation. 
We believe that next-token prediction is not only viable but also advantageous in the quest for general multimodal intelligence, bringing us closer to the realization of artificial general intelligence.

\section*{Contributors and Acknowledgements}

\subsection*{Project Lead}
\vspace{-1em}
Xinlong Wang

\subsection*{Contributors}
\vspace{-1em}
* indicates core contributors with equal contributions.

Xiaosong Zhang*, Zhengxiong Luo*, Quan Sun*, Yufeng Cui*, Jinsheng Wang*, Fan Zhang*, Yueze Wang*, Zhen Li*, Qiying Yu, Yingli Zhao, Yulong Ao, Xuebin Min, Tao Li, Boya Wu, Bo Zhao, Bowen Zhang, Liangdong Wang, Guang Liu, Zheqi He, Xi Yang, Jingjing Liu

\subsection*{Senior Leads}
\vspace{-1em}
Zhongyuan Wang, Yonghua Lin, Tiejun Huang

\subsection*{Acknowledgements}
\vspace{-1em}
We thank Xuan Liu, Shaokai Nie, Quanyue Ma, Hua Zhou, Yance Jiao, Liao Zhang, Mengyu Lu, Yiwen Shao, Yaohui Chen, Donglin Hao, Mengsi Lv, Teng Dai, Jiakang Liu for their support for the Emu3 project.

{
\bibliographystyle{plain}
\bibliography{ref}
}

\clearpage

\appendix
\section{Dataset Details}

\subsection{Video Dataset} 
\begin{figure}[ht]
    \centering
    \begin{minipage}{0.45\textwidth}
        \includegraphics[width=\linewidth]{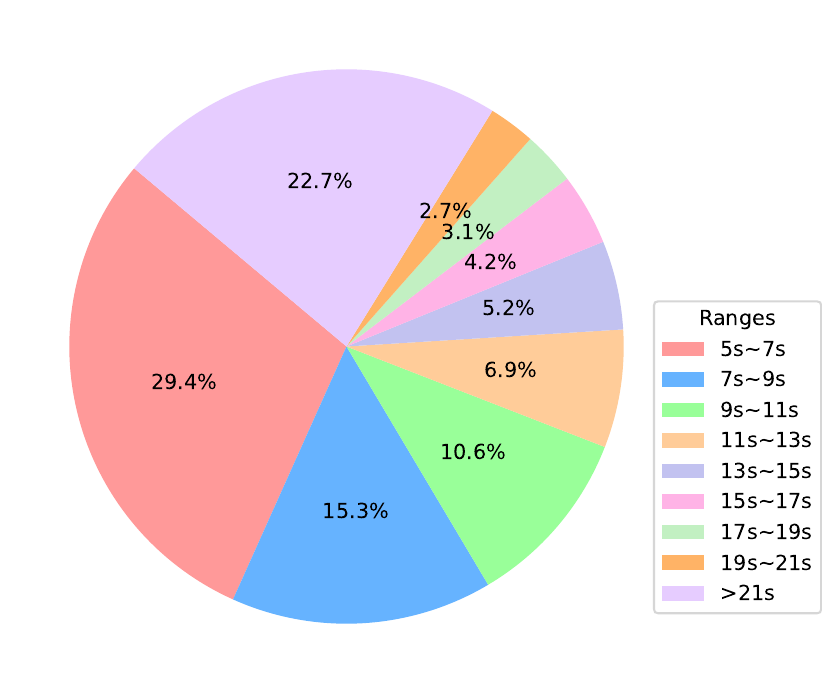}
    \caption{Duration distribution.}
    \label{fig:distr_video_len}
    \end{minipage}
    \begin{minipage}{0.45\textwidth}
        \includegraphics[width=\linewidth]{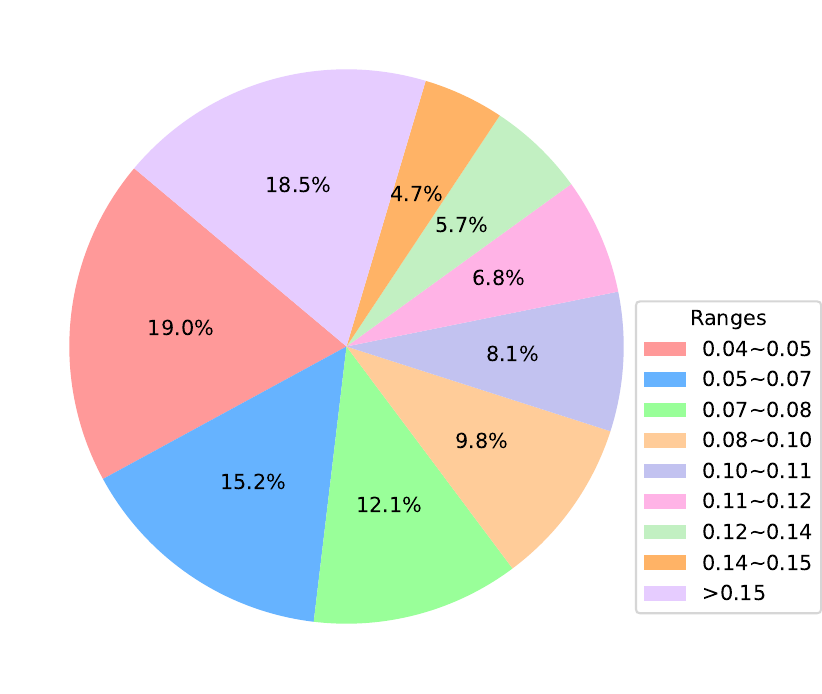}
    \caption{Flow score distribution.}
    \label{fig:distr_flow_score}
    \end{minipage}
\end{figure}

We analyze the distribution of the remaining clips. The duration distribution of the remaining clips is shown in~\cref{fig:distr_video_len}. The flow score distribution of filtered clips is shown in~\cref{fig:distr_flow_score}.

\section{Evaluation Details}

\subsection{Image Generation}
\label{sec:Image-eval}

For all T2I evaluations, we set Top-k to 16,384 and Top-p to 1.0 for image generation. The output resolution for \Ours is 512 x 512. The output resolution for \OursDPO is 720 x 720.

\textbf{Results on MSCOCO 30K.} We present zero-shot CLIP score and FID of \Ours and \OursDPO on MSCOCO 30K in~\cref{tab:text2image_fid}. Following ~\cite{sun2023emu}, we randomly sample 30k prompts from the validation set and calculate the zero-shot FID~\cite{heusel2017gans}. We employ CLIP-ViT-B~\cite{radford2021learning} to calculate the CLIP-T score to assess prompt-following ability. Additionally, we utilize CLIP-ViT-L~\cite{radford2021learning} to compute the CLIP-I score for measuring image similarity. For the DALL-E3 and DALL-E2, CLIP-T score is calculated on 4,096 samples. We adopt classifier-free guidance~\cite{ho2022classifier} for better generation quality. The guidance scale is set to 5.0. The results of other methods in the MSCOCO 30K are sourced from ~\cite{sun2023emu, show-o, zhou2024transfusion}

\textbf{Results on GenEval.}  Following SD3~\cite{esser2024scaling}, we evaluate text-to-image generation capability of \Ours on the GenEval benchmark~\cite{ghosh2024geneval}. We present the scores for the GenEval benchmark in ~\cref{tab:genevl} across six dimensions including “Single Object”, “Two Objects”, “Counting”, “Colors”, “Position”, “Color Attribute”. We generate 4 images for each prompt with a guidance scale of 5.5. Following with Dalle-3, we also report our evaluation results utilizing GPT4-V as a rewriter. The results of other methods in the GenEval are sourced from ~\cite{ghosh2024geneval, show-o, zhou2024transfusion, esser2024scaling}. 

\textbf{Results on T2I CompBench.} Folloing the Dalle-3~\cite{betker2023dalle3}, we report the scores of color binding, shape binding and texture binding in ~\cref{tab:genevl}. We use the BLIP-VQA model to evaluate these results. We generate 10 images for each prompt with a guidance scale of 5.0. The results of other methods in the T2I CompBench are sourced from ~\cite{betker2023dalle3, feng2024ranni, chen2023pixart}

\begin{table}[ht]
    \centering
    \resizebox{\linewidth}{!}{
        \begin{tabular}{l|ccccccc|ccc}
            \toprule
                    & \multicolumn{7}{c}{GenEval}  &  \multicolumn{3}{c}{T2I-CompBench}   \\
            Method  & Overall  & Single Obj. &  Two Obj. & Counting  & Colors & Position & Color Attri.  &  Color  & Shape & Texture  \\
            \midrule
            \multicolumn{9}{l}{\textit{Diffusion-based}} \\
            \midrule
            DALL-E 2  ~\cite{ramesh2022hierarchical} & 0.52 & 0.94 & 0.66 & 0.49 & 0.77 & 0.10 & 0.19 & 0.5750 & 0.5464 & 0.6374  \\
            SDv1.5~\cite{rombach2022high}            & 0.43 & 0.97 & 0.38 & 0.35 & 0.76 & 0.04 & 0.06 & 0.3730 & 0.3646 & 0.4219  \\
            SDv2.1~\cite{rombach2022high}            & 0.50 & 0.98 & 0.51 & 0.44 & 0.85 & 0.07 & 0.17    & 0.5694 & 0.4495 &  0.4982  \\
            SDXL~\cite{podell2023sdxl}               & 0.55 & 0.98 & 0.74 & 0.39 & 0.85 & 0.15 & 0.23    & 0.6369 & 0.5408 & 0.5637 \\
            PixArt-alpha~\cite{chen2023pixart}       & 0.48 & 0.98 & 0.50 & 0.44 & 0.80 & 0.08 & 0.07    & 0.6886  & 0.5582 & 0.7044  \\
            DALL-E 3 ~\cite{betker2023dalle3}   & 0.67  & 0.96 & 0.87 & 0.47 & 0.83 & 0.43 & 0.45    & 0.8110 & 0.6750 & 0.8070   \\ 
            SD3~\cite{esser2024scaling} &  0.74      & 0.99 & 0.94 & 0.72 & 0.89 & 0.33 & 0.60  & - & - & -  \\

            \midrule
            \multicolumn{9}{l}{\textit{Autoregressive meets diffusion}} \\
            \midrule
            Show-o~\cite{show-o}                     & 0.53 & 0.95 & 0.52 & 0.49 &  0.82 & 0.11 & 0.28 & - & - & - \\
            Transfusion~\cite{zhou2024transfusion}   & 0.63 & - & - & - & - & - & - & - & - & - \\

            \midrule
            \multicolumn{9}{l}{\textit{Autoregressive-based}} \\
            \midrule
            Chameleon~\cite{team2024chameleon}      & 0.39 & - & - & -  & - & - & - & - & - & - \\
            LlamaGen~\cite{sun2024autoregressive}   & 0.32 & 0.71 & 0.34 & 0.21  &  0.58 &  0.07 & 0.04 & - & - & - \\

            \Ours                                          & 0.54 & 0.98 & 0.71 & 0.34 & 0.81 & 0.17 & 0.21 & 0.6107 & 0.4734 & 0.6178  \\
            \hspace{0.3em} + \hspace{0.1em} Rewriter       & 0.66 & 0.99 & 0.81 & 0.42 & 0.80 & 0.49 & 0.45 & 0.7913 & 0.5846 & 0.7422  \\
            \OursDPO                                       & 0.52 & 0.98 & 0.69 & 0.33 & 0.78 & 0.15 & 0.16 & 0.5514 & 0.4641 & 0.5476  \\
            \hspace{0.3em} + \hspace{0.1em} Rewriter       & 0.64 & 0.99 & 0.76 & 0.38 & 0.85 & 0.45 & 0.40 & 0.7544 & 0.5706 & 0.7164  \\

            \bottomrule
        \end{tabular}
    }
    \vspace{0.5em}
    \caption{\textbf{Comparison with state-of-the-art models on GenEval and T2I CompBench.} Obj.: Object. Attri.: Attribute.}
    \label{tab:genevl}
\end{table}

\textbf{Results on DPG-bench.} To assess the ability to follow dense text, we compared our models with state-of-the-art (SoTA) diffusion models on the DPG-Bench, which provides longer prompts containing more detailed information for evaluation. We measured DPG-bench follows~\cite{hu2024ella} shown in the ~\cref{tab:dpg} , and our model achieved an overall score of 81.60, which is higher than SDXL and PixArt-alpha, and is comparable to the results of Dalle-3. We utilized mPLUG-large model to evaluate the generated images according to the designated questions. The results of other methods in the DPG-Benchmark are sourced from ~\cite{hu2024ella, liu2024playground}. We generate 4 images for each prompt with guidance scale is 5.0.

\begin{table*}[t]
\centering
\begin{tabular}{lcccccc}
\toprule 
    Method & Overvall & Global & Entity & Attribute & Relation & Other \\ 
    \midrule
    \multicolumn{7}{l}{\textit{Diffusion-based}} \\
    \midrule
    SDv1.5~\cite{rombach2022high} & 63.18 & 74.63 & 74.23 & 75.39 & 73.49 & 67.81 \\
    SDXL~\cite{podell2023sdxl} & 74.65 & 83.27 & 82.43 & 80.91 & 86.76 & 80.41 \\
    PixArt-alpha~\cite{chen2023pixart} & 71.11 & 74.97 & 79.32 & 78.60 &  82.57 &  76.96 \\
    Playground v2.5~\cite{li2024playground} & 75.47 & 83.06 & 82.59 & 81.20 & 84.08 & 83.50 \\
    Lumina-Next~\cite{zhuo2024luminanextmakingluminat2xstronger} & 74.63 & 82.82 & 88.65 & 86.44 & 80.53 & 81.82 \\
    Hunyuan-DiT~\cite{li2024hunyuan} & 78.87 & 84.59 & 80.59 & 88.01 & 74.36 & 86.41 \\
    PixArt-Sigma~\cite{chen2024pixartsigmaweaktostrongtrainingdiffusion} & 80.54 & 86.89 & 82.89 & 88.94 & 86.59 & 87.68 \\
    DALLE 3~\cite{betker2023dalle3} & 83.50 & 90.97 & 89.61 & 88.39 & 90.58 & 89.83 \\
    SD3-Medium~\cite{esser2024scaling} & 84.08 & 87.90 & 91.01 & 88.83 & 80.70 & 88.68 \\
    Playground v3~\cite{liu2024playground} & 87.04 & 91.94 & 85.71 & 90.90 & 90.00 & 92.72 \\
    \midrule
    \multicolumn{7}{l}{\textit{Autoregressive-based}} \\
    \midrule
    \Ours    & 80.60 & 85.21 & 86.68 & 86.84 & 90.22 & 83.15 \\
    \OursDPO & 81.60 & 87.54 & 87.17 & 86.33 & 90.61 & 89.75 \\
\bottomrule
\end{tabular}
\caption{\textbf{Comparison with state-of-the-art models on DPG-bench}}
\label{tab:dpg}
\end{table*}

\subsection{Post Processing}\label{sec:postprocessing}

To further improve the temporal consistency and visual quality, we apply stabilization and super resolution techniques to the generated videos. Video evaluation is also conducted on the processed videos. Specifically, we train specialized models for these two tasks.

\textbf{Video Stabilization.}
We train the video stabilization model based on the temporal VAE of stable video diffusion~\cite{blattmann2023stable}. The model is trained on our curated video data with a combined objective comprising L1 loss, LPIPS perceptual loss~\cite{lpips}, GAN loss, and KL penalty~\cite{kingma2013auto,rezende2014stochastic}. A training data pair consists of an autoencoded video clip output from our tokenizer and the groundtruth video clip, both having dimensions of $16 \times 256 \times 256$.

\textbf{Super-Resolution.}
We implement a spatial-temporal unet model for super-resolution task, capable of upsampling any image or video clip by a factor of 4. We adopt the BlurPool~\cite{zhang2019making} for downsample operations and sub-pixel~\cite{shi2016real} for upsample operations. The model is trained on random crops of $8 \times 256 \times 256$ from part of our curated videos, which have a resolution greater than $1024 \times 1024$, with a combined loss of L2 loss, LPIPS perceptual loss~\cite{lpips}, and GAN loss.

\clearpage

\section{Qualitative Examples for Multimodal Understanding}

\begin{table}[ht]
    \centering
    \begin{minipage}{\linewidth}
        \small
        \begin{tabular}{p{2.0cm} p{11.0cm} }
        \toprule
         \multicolumn{2}{l}{\bf Example}  \\
        \midrule
        &  \includegraphics[height=3.5cm]{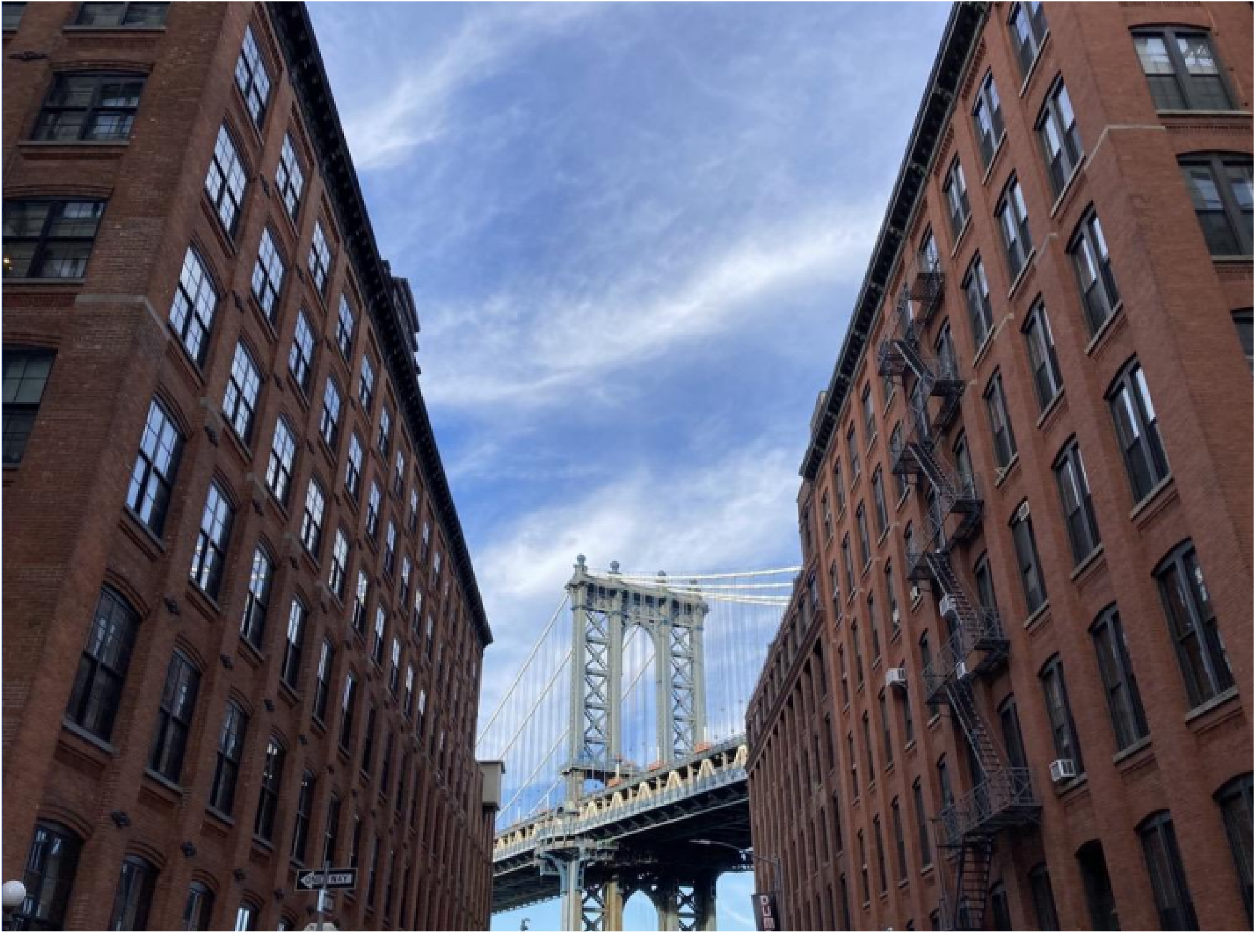} \\
        User & Describe the landmark. \\
        \midrule
        Emu3 & {\small The image showcases the Brooklyn Bridge, a prominent suspension bridge in New York City, recognized by its distinctive towers and the web of cables that support it. This iconic structure connects the boroughs of Brooklyn and Manhattan.} \\
        \bottomrule
        \end{tabular}
        
    \end{minipage}
    \vspace{0.1em}
    \vfill
    \begin{minipage}{\linewidth}
        \small
        \begin{tabular}{p{2.0cm} p{11.0cm} }
        \toprule
         \multicolumn{2}{l}{\bf Example }  \\
        \midrule
        &  \includegraphics[height=1.03cm]{figs/i2t/OCRBench_977.png} \\
        User & Please write out the expression of the formula in the image using LaTeX format. \\
        \midrule
        Emu3 & {\small 
        $S_{\Delta } = \frac { \sqrt { 2 } \cdot \sqrt { 7 } } { 2 } = \frac { \sqrt { 1 4 } } { 2 }$
        } \\
        \bottomrule
        \end{tabular}
    \end{minipage}

    \vspace{0.1em}
    \vfill
    \begin{minipage}{\linewidth}
        \small

        \begin{tabular}{p{2.0cm} p{11.0cm} }
        \toprule
         \multicolumn{2}{l}{\bf Example }  \\
        \midrule
        &  \includegraphics[height=5cm]{figs/i2t/ai2d_2.png} \\
        User & According to the given food chain what would happen if phytoplankton decreases?
        
        A. Seal population will become extinct
        
        B. Fish population would decrease.
        
        C. Whale population would decrease.
        
        D. Penguin population would increase.
        
        Answer with the option's letter from the given choices directly. \\
        \midrule
        Emu3 & {\small B } \\
        \bottomrule
        \end{tabular}
    \end{minipage}

\end{table}

\begin{table}[ht]
    \centering
    \begin{minipage}{\linewidth}
        \small
        \begin{tabular}{p{2.0cm} p{11.0cm} }
        \toprule
         \multicolumn{2}{l}{\bf Example }  \\
        \midrule
        &  \includegraphics[height=6cm]{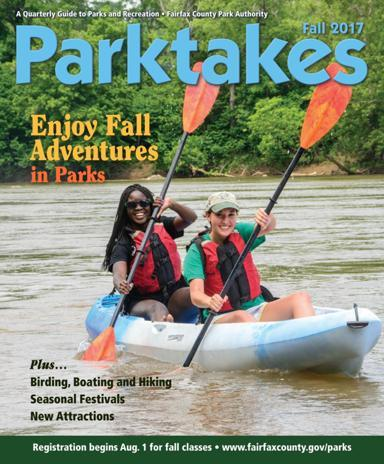} \\
        User & Please describe the image in detail \\
        \midrule
        Emu3 & {\small 
        The image is the cover of a magazine titled "Parktakes," which is the Fall 2017 issue. The cover features two individuals in a white kayak, paddling on a body of water. The kayak is positioned diagonally, with the two individuals seated towards the front. Both individuals are wearing life jackets, which are red with black straps, and they are holding paddles with orange blades. The person on the left is a woman with long dark hair, wearing sunglasses and a black top. The person on the right is a man with short hair, wearing a green shirt and a light-colored cap. They both appear to be smiling and enjoying their time on the water.
        
        The background shows a natural setting with trees and greenery, indicating that they are likely in a park or natural reserve. The water is calm, and the sky is not visible, suggesting a serene and peaceful environment.
        
        The magazine cover includes several text elements:
        
        1. **Title**: "Parktakes" is prominently displayed at the top in large, bold, blue letters.
        
        2. **Fall 2017 Issue**: The issue is marked in the top right corner in a smaller, white font.
        
        3. **Main Feature**: "Enjoy Fall Adventures in Parks" is written in a large, bold, orange font, emphasizing the theme of outdoor activities.
        
        4. **Additional Features**: The cover lists several topics covered in the magazine:
        
           - "Birding, Boating, and Hiking"
           
           - "Seasonal Festivals"
           
           - "New Attractions"
           
        5. **Registration Information**: At the bottom, there is a call to action to "Registration begins Aug. 1 for full classes" and a website link: "www.fairfaxcounty.gov/parks."
        
        Analysis and Description
        
        The cover of "Parktakes" effectively conveys the theme of outdoor recreational activities, particularly in parks. The image of the two individuals in a kayak suggests that the magazine features content related to kayaking, canoeing, and other water sports. The inclusion of life jackets indicates a focus on safety and preparedness for water activities. The mention of seasonal festivals and new attractions implies that the magazine includes information on various events and attractions within the parks.
        
        The cover also highlights the magazine's commitment to providing a comprehensive guide to outdoor activities, with a clear emphasis on safety and accessibility. The registration information suggests that the magazine is a resource for planning and participating in various outdoor activities, likely targeting a wide audience interested in parks and recreational activities.
        
        In summary, the magazine cover is designed to attract readers interested in outdoor adventures, safety, and seasonal events within parks. The use of a kayak and life jackets emphasizes the magazine's focus on water sports and safety, while the inclusion of various features and registration information suggests a well-rounded resource for park visitors. } \\
        \bottomrule
        \end{tabular}
    \end{minipage}
\end{table}

\end{document}